\documentclass{ijcaArticle}
\ijcaVolume{VV}
\ijcaNumber{N}
\ijcaYear{YYYY}
\ijcaMonth{Month}

\usepackage[ruled]{algorithm2e}
\usepackage{graphicx}
\usepackage{amsmath}
\usepackage{amsfonts}
\usepackage{booktabs}
\usepackage{tabularx}
\usepackage{appendix}
\usepackage{placeins}

\usepackage{float}
\usepackage{tikz, pgfplots}
\usetikzlibrary{spy}
\usepgfplotslibrary{groupplots}
\pgfplotsset{compat=newest}
\usepackage{subcaption}

\usepackage{biblatex}
\newcommand{\citet}{\cite}

\ijcaVolume{*}
\ijcaNumber{*}
\ijcaYear{2012}
\ijcaMonth{--------}
\begin{document}

\title{Pruning in the Face of Adversaries}

\author{ 
   \large Florian Merkle \\[-3pt]
   \normalsize Management Center Innsbruck  \\[-3pt]
    \normalsize Digital Business and Software Engineering \\[-3pt]
    \normalsize Innsbruck \\[-3pt]
    \normalsize florian.merkle@mci.edu \\[-3pt]
  \and
   \large Maximilian Samsinger \\[-3pt]
   \normalsize Management Center Innsbruck  \\[-3pt]
    \normalsize Digital Business and Software Engineering \\[-3pt]
    \normalsize Innsbruck \\[-3pt]
    \normalsize maximilian.samsinger@mci.edu \\[-3pt]
\and
   \large Pascal Schöttle \\[-3pt]
   \normalsize Management Center Innsbruck  \\[-3pt]
    \normalsize Digital Business and Software Engineering \\[-3pt]
    \normalsize Innsbruck \\[-3pt]
    \normalsize pascal.schoettle@mci.edu \\[-3pt]
}


\maketitle

\begin{abstract} 
  The vulnerability of deep neural networks against adversarial examples -- inputs with small imperceptible perturbations -- 
    has gained a lot of attention in the research community recently.
    Simultaneously, the number of parameters of state-of-the-art deep learning models has been growing massively, with implications on the memory and computational resources required to train and deploy such models. One approach to control the size of neural networks is retrospectively reducing the number of parameters, so-called neural network pruning.
    
    Available research on the impact of neural network pruning on the adversarial robustness is fragmentary and often does not adhere to established principles of robustness evaluation.
    We close this gap by evaluating the robustness of pruned models against $L_0$, $L_2$ and $L_\infty$ attacks for a wide range of attack strengths, several architectures, data sets, pruning methods, and compression rates.
    
    Our results confirm that neural network pruning and adversarial robustness are not mutually exclusive. Instead, sweet spots can be found that are favorable in terms of model size and adversarial robustness. 
    Furthermore, we extend our analysis to situations that incorporate additional assumptions on the adversarial scenario and show that depending on the situation, different strategies are optimal.
\end{abstract}

\section{Introduction}
Modern deep neural networks (DNNs) are increasingly able to solve sophisticated tasks from computer vision to natural language processing and beyond. This has substantial implications for mankind and thus, we expect DNNs to behave as intended.
However, \cite{szegedy2013intriguing} found that adversarial examples, minimally perturbed input samples, can fool DNNs into misclassification. 

While much of the current research on adversarial robustness
is conducted in an artificial, virtual setting, some work has shown that adversarial machine learning is applicable to real-world scenarios, such as road sign classification \cite{eykholt2018robust}, fooling voice-assistants\footnote{https://nicholas.carlini.com/code/audio\_adversarial\_examples/} or face-recognition software with adversarial patterns on eyeglass frames \cite{sharif2016accessorize}.

Although the availability of computational resources drove the recent progress in the field of deep learning, there are many applications where resources are scarce. Deep learning applications deployed on IoT or mobile devices and real-time applications heavily rely on the resource optimization. 
Previous work \cite{madry2018towards} has suggested that there is a direct relation between a model's capacity, i.e., the number of parameters of a model, and its respective adversarial robustness.
Bigger models are more memory and computationally intensive, and some domains impose restrictions on the resources a model may use. The resources might be bounded by the available hardware or economic aspects. Neural Network (NN) pruning~\cite{lecun1990optimal} slims down a model's size before deployment, decreasing memory usage and increasing computational efficiency for inference. 

Recently, some research on examining the adversarial robustness of pruned NNs has started to emerge. However, much of this work is fragmentary, lacks a clear threat model, suffers from an inadequate choice of attacks, or does not adhere to other principles of a rigorous robustness evaluation as described by \citet{carlini2019evaluating}. Consequently, existing literature does not provide clear results on the impact of NN pruning on the adversarial robustness.

We conduct an exhaustive study which covers the most relevant attacks, pruning methods, architectures, and data sets. Hereby, we confirm evidence from previous work that NN pruning does not necessarily impact a model's adversarial robustness negatively for various combinations of factors.
We show that NN pruning provides a particular space for optimal strategies, balancing clean and robust accuracy.

The remainder of this paper is organized as follows: Section~\ref{sec:relwork} introduces the necessary theoretical foundation and covers the current state of the research on the adversarial robustness of pruned NNs. Section~\ref{sec:experimental-setup} presents the design of our experiments. Specifically, we introduce and elaborate on our choice of architectures, attacks, and pruning methods, before we present the experimental results and discuss their relevance on a defender's pruning strategy in Section~\ref{sec:results}. Finally, Section~\ref{sec:conclusion} concludes this paper and highlights implications of our work for further research and real-world-scenarios.

\section{Related Work}
\label{sec:relwork}
In this section, we present related work on NN pruning, adversarial machine learning, and the current state-of-the-art in the combination of those fields.
\subsection{Neural Network Pruning}
Network pruning refers to the deletion of parameters of a DNN. Modern NNs are typically over-parameterized for the task at hand, leading to extensive redundancy in the model~\cite{liu2018rethinking}.  
The goal of pruning is to reduce storage, memory usage, and computational resources. 
Interestingly, it has been shown that it is possible, by carefully selecting the parameters to be removed, to not only reduce the resource requirements of a model without suffering performance losses but instead to increase the accuracy simultaneously~\cite{frankle2018lottery,han2015learning}.

\noindent
Pruning approaches can be described on five dimensions:

The \textbf{structure} describes the granularity of a method.
The unstructured approach prunes single weights \cite{lecun1990optimal,blalock2020state} while structured pruning removes entire \textit{parts}, such as kernels and filters \cite{he2018soft}, or even whole residual blocks \cite{huang2018data}. As the first approach produces sparse matrices of the same size as the unpruned network, dedicated hardware is necessary to accomplish optimizations.

The \textbf{selection criterion} defines how to select the parameters to be pruned. Many approaches have been proposed: Based on their absolute values \cite{han2015learning}, the gradients \cite{blalock2020state}, on the Hessian matrix of the loss function \cite{lecun1990optimal}, or the $L_2$ norm of a structure \cite{he2018soft}.
Network pruning can also be incorporated into the learning procedure \cite{huang2018data} or formulated as its own optimization problem \cite{zhang2018systematic}. Finally, random pruning can serve as a baseline and sanity check \cite{blalock2020state,frankle2018lottery}. 

The \textbf{scope} determines whether the selection process is performed locally \cite{he2018soft}, where each layer is pruned separately, or globally where all weights are considered simultaneously for the selection process.

\textbf{Scheduling} determines when pruning is conducted. Most methods, e.g. \cite{han2015learning}, apply pruning after training.
The network is either pruned in one step to the desired compression rate~\cite{liu2018rethinking} 
 or as an iterative process of pruning and consequent training \cite{han2015learning,gale2019state}.

\textbf{Fine tuning} refers to the training phase after pruning is applied. Traditionally fine tuning is conducted with the pre-pruned weight values \cite{han2015learning}, but recent work explores differences when re-initializing the weights with its initial random values~\cite{frankle2018lottery} or a set of new random values~\cite{liu2018rethinking}.

\subsection{Adversarial Machine Learning}
Research on adversarial machine learning started in 2004 when it was first explored that spam filters utilizing linear classifiers can be fooled by small changes in the initial email that do not negatively affect the readability of the message but lead to misclassification \cite{dalvi2004adversarial}. 

In 2013 \citet{szegedy2013intriguing} showed that DNNs are just as prone to adversarial examples as other machine learning algorithms. They confirmed their findings for several architectures and datasets.

Formally, an \textbf{Adversarial Example} can be described as follows: A classifier is a function $C(x) = y$  that takes an input $x$ and yields a class $y$. If a perturbation $\delta$ is added to $x$, such that the manipulated input $x + \delta = \tilde{x}$ leads to a classification different from the original value $C(\tilde{x}) \neq C(x)$, it is labeled an adversarial example.
Usually, distance metrics are used to quantify the difference between $x$ and $\tilde{x}$.
In image classification, the most important metrics are~\cite{carlini2017towards}:
\begin{itemize}
  \item The $L_0$ distance:  counts the number of elements $i$ where $x_i \ne \tilde{x}_{i}$, i. e. the total number of pixels changed.
  \item The $L_2$ norm: $||x - \tilde{x}||_2 = ( \sum_{i=1}^n|x_i - \tilde{x}_i|^2)^\frac{1}{2}$
  \item The $L_{\infty}$ norm: $||x - \tilde{x}||_\infty = \textrm{max}\{|x_1 - \tilde{x}_1|,..., |x_n - \tilde{x}_n|\}$, i. e. the maximum distance. 
\end{itemize}

\subsubsection{Threat Model} \label{sec:Threat Model} 
Security evaluations should always state a precise threat model, i. e., the assumptions about an adversary's goals, knowledge, and capabilities~\cite{carlini2019evaluating}.

\textbf{Adversary Goals}
In adversarial machine learning, the adversary's goal can either be an untargeted attacks where $C(x) \neq C(\tilde{x})$ or a targeted attacks $C(\tilde{x}) = t$, where $t$ is a defined target class.

\textbf{Adversary Capability}
It makes sense to restrict the capabilities of an adversary. Without restrictions she would be able to manipulate the input pipeline, evade the model at training time, change the semantics of an input image, or even make hard changes on the model's weights.

Most works impose constraints so that an adversary can make only small changes to an input. A valid adversarial example $\tilde{x}$ would fulfill $D(x, \tilde{x}) \leq \epsilon$, where $\epsilon$ is the upper boundary of the allowed alteration and $D$ is a similarity metric. $\epsilon$ can be interpreted as the strength of an attack as higher values yield a strictly greater accuracy loss.

\textbf{Adversary knowledge}
An adversary has a certain level of knowledge of the targeted model regarding the training data, the optimization algorithm, the loss function, hyperparameters, the DNN architecture, or the learned parameters. If an adversary has access to all this information, the setting is labeled a white-box attack. This allows a worst-case scenario evaluation of the examined model. In the black-box setting, the adversary has no knowledge of the model, but might have (limited) access to the model to retrieve information~\cite{carlini2019evaluating}. 

\subsubsection{Attacks}
\label{sec:attacks_theory}
Recently, a wide variety of attack algorithms emerged from the research community. Adversarial attacks can be divided into gradient-based attacks which require full access to the model and its weights and black-box attacks that rely either on meaningful model outputs such as logits or probabilities or solely on its final decision~\cite{rauber2017foolbox}. We introduce the algorithms relevant for this work in detail in Section~\ref{sec:attacks}.

\subsubsection{Countermeasures}
Multiple approaches to defend against adversarial attacks have been proposed but only few have so far remained unbroken \cite{tramer2020adaptive}.

\textbf{Adversarial (re)training}, e. g.~\cite{madry2018towards}, enhances a model's robustness by presenting the model adversarial examples during training. 
Adversarial (re)training reduces the clean accuracy, and due to the necessary additional backward passes, induces high computational costs. 

Another direction of research aims to achieve \textbf{certifiable robustness}. 
\textit{Randomized smoothing} transforms the problem of classifying under adversarial perturbations into the simpler problem of classifying under random noise. Cohen et al.~\cite{cohen2019certified} guarantee a certain level of accuracy under any norm-bounded attack up to a specific attack strength by inducing Gaussian noise at training time, and an additional \textit{smoothed} classifier. 
Certifiable robustness induces computational complexity and the certified robustness is only a fraction of the empirical robustness gained from adversarial (re)training.

\subsection{Robustness of Pruned Networks}
A limited body of research is available on the effect of network pruning on adversarial robustness. However, most of the work focuses on successive or concurrent adversarial training and NN pruning.

One string of research aims to unify the act of NN pruning and adversarial training into a single framework. \citet{ye2019adversarial} evaluate the adversarial robustness of an ADMM-based pruning method, implemented for weight, column, and filter-based pruning on the VGG-16~\cite{neklyudov2017structured} and ResNet-18~\cite{he2016deep} architecture. They apply two $L_\infty$ attacks \cite{madry2018towards,carlini2017towards} and find that for concurrent adversarial training and pruning, the robustness decreases the higher the compression rate is. However, they show that a bigger model that is pruned retrospectively is more robust than an unpruned model with a similar parameter count.

Another string of work \cite{sehwag2020hydra} proposes to align network training and pruning and to make the pruning process aware of the training objective, which can be defined as empirical or verified adversarial robustness. The pruning problem is solved with SGD and assigns an importance score to each weight. They evaluate robustness against PGD \cite{madry2018towards}, and the auto-attack ensemble \cite{croce2020reliable} and claim state-of-the-art clean and robust accuracy.

Finally, \citet{guo2018sparse} examine the robustness of a fully connected LeNet 300-100 \cite{lecun1998gradient}, a LeNet-5 \cite{lecun1998gradient}, a VGG-like network \cite{neklyudov2017structured} and a ResNet \cite{he2016deep}. They applied two $L_{\infty}$ attacks \cite{goodfellow2014explaining,tramer2018ensemble} and two $L_2$ attacks \cite{carlini2017towards,moosavi2016deepfool} on naturally trained models. The authors find that the sparse DNNs are consistently more robust to FGSM attacks than their respective dense models. As they only evaluated the robustness for one perturbation budget and applied only one unstructured-magnitude pruning approach, no general robustness assertion can be drawn.

\section{Experimental Setup}\label{sec:experimental-setup}

In this section, we elaborate and give a rationale for the design of our experiments. While we are aware that NN pruning is no defense mechanism, we adopt the principles of rigorous robustness evaluation as proposed by Carlini et al.~\cite{carlini2019evaluating}, where applicable. In total, we evaluate the adversarial robustness of NNs for a combination of four architectures, three attack methods, four perturbation budgets, nine pruning methods, and seven compression rates, yielding a total of 3\,048 data points.

\subsection{Threat Model}
With regards to the threat model introduced above, we follow the recommendation of \citet{carlini2019evaluating} and model the strongest adversary possible. As such, the adversary's \textit{goal} is untargeted misclassification, and we assume the white-box scenario in which the adversary has perfect \textit{knowledge}. That is: access to the model architecture, the data used for learning, its weights, and gradients. We grant the adversary different levels and forms of \textit{capabilities}. 
We have chosen three attacks, one for each of the $L_0$, $L_2$, and $L_\infty$ distance metrics, and perform them in various strengths. Furthermore we consider an adaptive adversary, so when evaluating the robustness, we use the exact same -- possibly pruned -- model that is under attack to craft the adversarial examples.

\subsection{Adversarial Attacks}\label{sec:attacks}
As gradient masking is not an issue in the pruning setting, we focus our experiments on gradient-based attacks.
We partly follow the recommendations of \citet{carlini2019evaluating}, however as we expect similar results for $L_1$ and $L_2$ distortion, we opt to drop $L_1$ and instead analyze the robustness under an $L_0$ attack.
We considered the $L_0$ attacks proposed by Carlini and Wagner~\cite{carlini2017towards} and \citet{brendel2019accurate}. Preliminary experiments on our smallest model showed that both attacks are viable choices to evaluate the $L_0$ robustness. We choose the Brendel\&Bethge attack~\cite{brendel2019accurate} as it is significantly less computationally expensive.
Other than that, we evaluate the attacks proposed by \cite{carlini2019evaluating}. This gives us the following array of attacks: 
\begin{itemize}
    \item Brendel\&Bethge \cite{brendel2019accurate} for $L_0$ perturbations
    \item Carlini\&Wagner \cite{carlini2017towards} for $L_2$ perturbations
    \item PGD \cite{madry2018towards} for $L_{\infty}$ perturbations
\end{itemize}
We apply each attack with a set of four $\epsilon$-values, which we have chosen such that the weakest attack does not have any impact on the \textit{unpruned} model and the strongest attack fools the same model for more than 50\% of the test inputs. The values differ with regards to the specific architecture and the data set and are available in Appendix \ref{appendix-attacks}. Furthermore, upon acceptance of the paper, we will publish all our experiments on GitHub for reproducibility.

\textbf{Brendel\&Bethge (B\&B) $L_0$:} Unlike many common attacks, this attack does not start from a clean sample but instead uses a starting point that is adversarial, but potentially with a high distance to the clean image. In every iteration the attack solves a quadratic trust region minimization problem to find a perturbation $\delta^i$ for the step $i$, so that the $L_p$ distance between $x$ and the updated perturbed image $\tilde{x} ^i = \tilde{x} ^{i-1} + \delta^i$ is minimal, $||\delta_i||_2^2$ stays within the trust region around $x^{i-1}$ with a defined radius $r$, respects the box constraint for a valid image $0 \leq x^{i-1} \leq 1$, and the perturbed image $x^i$ lies on, or close to the decision boundary. 
Joining these constraints, the following optimization problem is formulated:

\begin{gather}
\label{eq:bb}\nonumber
    \min\limits_{\delta} \; ||x - \tilde{x}^{i-1} - \delta^i ||_p \;\text{s.t.} \\
    \quad 0 \leq \tilde{x}^{i-1} + \delta^k \leq 1 \; 
    \wedge 
    \; b^{i{\mathsf{T}}}\delta^i = adv(\tilde{x^{i-1}}) \; 
   \wedge 
    \; ||\delta^k||_2^2 \leq r
\end{gather}
$||\cdot||_p$ is the $L_p$ norm, in our case the $L_0$ distance, and $b^i$ is the normal vector of the decision boundary in the area around $x^{i-1}$. In order to find the boundary between the adversarial and the non-adversarial space, 
the authors introduce the adversarial criterion $adv(\tilde{x}) = \min_{t, t \neq y}(Z(\tilde{x})_y - Z(\tilde{x})_t)$  where $Z(\tilde{x}) \in \mathbb{R}^C$ are the logits of the model and $t$ is the second most probable class for untargeted attacks. The boundary is given by a hypersurface for which $adv(\tilde{x}) = 0$.

The derivative of the adversarial criterion denotes the direction of the boundary $b^i$ at a specific step $i$ for the point $\tilde{x}^{i-1}$: 
\begin{equation}
    b^i = \nabla_{\tilde{x}^{i-1}} \text{adv}(\tilde{x}^{i-1})
\end{equation}

\newcolumntype{Y}{>{\centering\arraybackslash}X}

\begin{table*}
\centering
\begin{tabularx}{.8\textwidth}{l  YYYYYYY}
 & \multicolumn{7}{c}{Compression Rate}\\
\cmidrule{2-8}
Pruning method &  1  &  2  &  4  &  8  &  16 &  32 &  64 \\
\cmidrule{1-8}
magnitude global filter & 85.37 & \textbf{86.37} & 83.96 & 79.55 & 66.62 & 50.76 & 37.92 \\
magnitude global kernel & 84.43 & 86.35 & \textbf{87.01} & 85.75 & 81.63 & 75.71 & 49.44 \\
magnitude global unstructured      & 85.57 & 86.21 & 86.58 & \textbf{86.82} & 86.01 & 84.60 & 82.36 \\[0.5ex]
magnitude local filter  & 85.79 & \textbf{86.66} & 84.14 & 80.38 & 76.91 & 58.88 & 42.38 \\
magnitude local kernel  & 85.14 & 86.08 & \textbf{86.14} & 84.93 & 79.32 & 53.57 & 48.03 \\
magnitude local unstructured       & 85.44 & 86.07 & \textbf{86.68} & 86.04 & 85.24 & 83.61 & 81.77 \\ [0.5ex]
random local filter     & \textbf{86.31} & 86.26 & 86.02 & 83.95 & 80.35 & 75.44 & 61.51 \\
random local kernel     & 85.22 & 86.67 & \textbf{87.33} & 86.11 & 81.96 & 39.99 & 64.12 \\
random local unstructured          & 84.96 & \textbf{87.05} & 86.85 & 86.79 & 86.20 & 76.69 & 48.79 \\[0.5ex]

\end{tabularx}
\caption[Clean accuracies for all pruning methods and compression rates]{Clean accuracies of the ResNet18 for all pruning methods and compression rates. Bold numbers indicate the highest clean accuracy per row.}
\label{tab:clean_accuracies}\end{table*}

So at every step, if $b^{i\mathsf{T}}\delta^i = adv(\tilde{x}^{i-1})$, the perturbed input $\tilde{x}^i$ moves along the boundary closer towards the clean input $x$. To solve the optimization problem, the authors propose the Nelder-Mead algorithm \cite{nelder1965simplex} for the $L_0$ metric on the dual of the initial problem formulation presented in Equation~\eqref{eq:bb}.

We sample our starting points directly from the data set, i.e., for every image to be attacked, we use another image that is assigned a different class.

\textbf{Carlini\&Wagner (C\&W) $L_2$:}
\citet{carlini2017towards} substitute the perturbation $\delta$ with $\frac{1}{2} (\tanh(w) + 1) - x$ to ensure the adversarial example is within the box constraint $0\leq x+\delta \leq 1$.
The C\&W $L_2$ attack algorithm than aims to find $w$ for a chosen target class $t$:
\begin{equation}
\min ||\frac{1}{2} (\tanh(w) + 1) - x||_2^2 + c * f(\frac{1}{2} (\tanh(w) + 1)
\end{equation}
where $f$ is defined as:
\begin{equation}
    f(\tilde{x}) = \max((\max\limits_{s\neq t}({Z(\tilde{x})_s} ) - Z(\tilde{x})_t, -\kappa)
\end{equation}
$s$ is the second most probable class, $c$ is a constant that controls which term is optimized first and $\kappa$ is a parameter that controls the confidence. We set $\kappa$ to 0, so that the target class should be only marginally more probable then the second most probable class.

\textbf{PGD $L_{\infty}$:}
Projected gradient descent (PGD) starts from a benign sample and iterates over the following equation, in which the inner function is the fast gradient sign method~\cite{madry2018towards}:

\begin{equation}
\tilde{x}_{i+1} = P( x_i + t \textrm{ sign } (\nabla_x L(\theta, x, y )))
\end{equation}

$t$ is the defined step size, and the projection $P$ back into the allowed set is realized by clipping the updated values such that $0\leq x+\delta \leq 1$ and the perturbation is less than the given perturbation budget $\epsilon$, i.e., $\delta \leq \epsilon$.

\subsection{Pruning Methods}
The selection of pruning methods is motivated by~\citet{paganini2020iterative} and~\citet{blalock2020state}. We implement unstructured and structured pruning. For the structured approach, we consider both kernel- and filter-wise pruning. When pruning structures, we only prune the convolutional layers. This leads to a slightly lower total sparsity of the network but no significant reduction of the theoretical speed-up.

Magnitude-based pruning methods are a reliable choice as it is widely adopted in current research, and it has been proven to yield competitive results in comparison with more sophisticated approaches \cite{blalock2020state}.
Additionally, we implement random pruning as a baseline and sanity check.
For magnitude-based pruning, we examine local and global pruning.
We renounce global random pruning as for a large enough number of structures, it will yield the same pruning masks as local random pruning. This leaves us with the following nine pruning methods:

\begin{itemize}
    \item Unstructured \{local magnitude$\vert$global magnitude$\vert$local random\}  pruning
    \item Kernel-wise \{local magnitude$\vert$global magnitude$\vert$local random\}  pruning
    \item Filter-wise \{local magnitude$\vert$global magnitude$\vert$local random\}  pruning
\end{itemize}

As \citet{han2015learning} have shown, magnitude-based methods yield better results when conducted iteratively. Thus we refrain from examining one-shot pruning and implement our pruning methods strictly with an iterative pruning schedule.
At each pruning step, the network is trained to convergence, and subsequently, half of the remaining weights are removed. For the choice of compression rates, we follow the recommendation of \citet{blalock2020state} to use the set ${2,4,8,16,32}$ and add 64 in order to attain more expressive curves. 
After the pruning procedure, the remaining weights are retained for retraining as this reduces the computational effort. \citet{paganini2020iterative} have shown that the results for weight retention \cite{han2015learning}, re-initialization \cite{liu2018rethinking} and rewinding \cite{frankle2018lottery} yield comparable results.

\subsection{Architectures and Data Sets}
We apply all pruning methods and attacks on four different architectures. We have chosen a five-layer convolutional neural network, referred to as CNN5 from here, and a VGG11-like architecture due to its simplicity. Additionally, we expand our experiments to ResNets with pre-activation residual blocks~\cite{he2016identity} with 8 and 18 layers, respectively.
We train and evaluate the CNN5 network on the MNIST data set \cite{lecun2010mnist}, the VGG11-like and the ResNet8 on the CIFAR10 data set \cite{krizhevsky2009learning} and the ResNet18 on the Imagenette data set \cite{howard2019imagenette}.

\begin{figure*}
    \centering
    \resizebox{\linewidth}{!}{\input{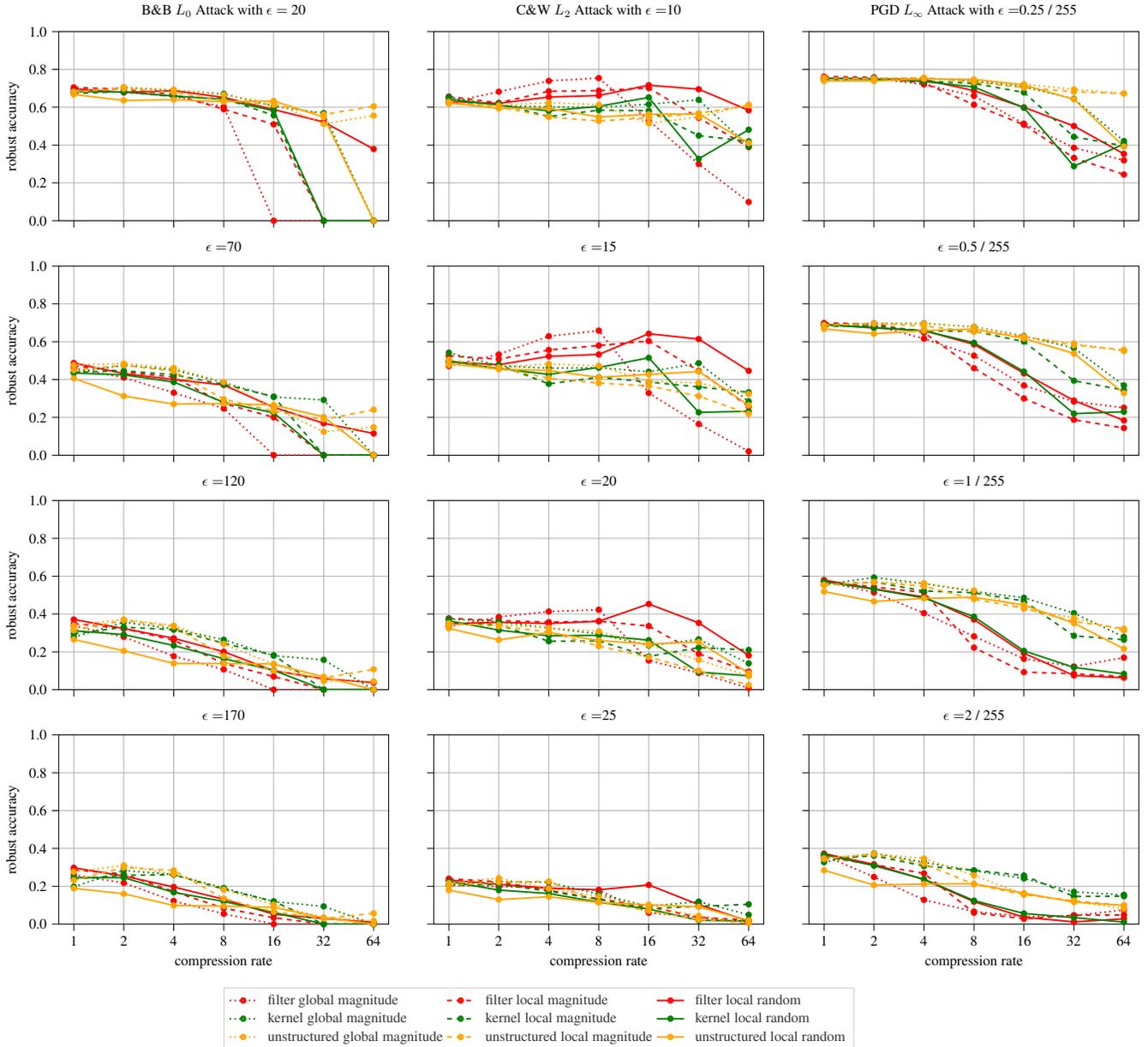}}
    \caption[]{Adversarial robustness of the ResNet18 models against the B\&B $L_0$ (\textbf{left} column), C\&W $L_2$ (\textbf{middle}) and PGD $L_\infty$ attacks (\textbf{right}) with increasing attack strengths from top to bottom. Each line in each plot depicts a pruning method over the compression rates from one to 64. (Best viewed in color.)}
    \label{fig:resnet18}
\end{figure*}

The CNN5 consists of two convolutional layers with a kernel size of five and three dense layers. We adapt the first and last layers of the VGG11-like and the ResNet8 to fit the CIFAR10 data set. On the VGG11-like architecture, we add batchnorm \cite{ioffe2015batch} layers for enhanced trainability. On the ResNet18, we adapt the last layer to match the ten classes of the Imagenette data set.

\subsection{Evaluation}
Our approach can be seen as a model with variables, dependent and independent, and constants. Adversarial robustness is the dependent variable. The compression rate and attack strength are independent variables. We measure the change in adversarial robustness on a selection of network architectures, pruning methods, and adversarial attacks. 

We define adversarial robustness as the accuracy of a model $f$ under attack, i.e. the fraction of all images in a data set of size $n$ where the model predicts the correct class for the perturbed image. 
\begin{equation}
    Acc_{robust} = \frac{\sum\limits_{i=1}^n f(\tilde{x}) = y}{n}
\end{equation}

\begin{table*}[t]
\centering
\begin{tabularx}{.8\textwidth}{lcYYYY}
 & \multicolumn{5}{c}{Compression Rate}\\
\cmidrule{2-6}
{Test Data}  &     1 & 2 &    4 &  8 &    16 
\\
\midrule
benign      & 85.57 & 86.21 (\textbf{0.64}) & 86.58 (\textbf{1.01}) &  86.82 (\textbf{1.25}) & 86.01 (\textbf{0.44}) 
\\[.5ex]
$l_0, \epsilon=20$    & 68.05 & 70.62 (\textbf{2.58}) & 69.30 (\textbf{1.25}) & 66.56 (-1.48) & 62.11 (-5.94) 
\\
$l_0, \epsilon=70$   & 47.66 & 48.44 (\textbf{0.78}) & 46.17 (-1.48) & 38.67 (-8.98) & 25.08 (-22.58)
\\
$l_0, \epsilon=120$  & 33.91 & 37.19 (\textbf{3.28}) & 33.19 (0.00) & 24.22 (-9.69) & 13.44 (-20.47) 
\\
$l_0, \epsilon=170$  & 27.66 & 31.02 (\textbf{3.36}) & 26.44 (-1.33) & 18.28 (-9.38) & 10.23 (-17.42) 
\\[.5ex]
$l_2, \epsilon=10$  & 61.88 & 60.94 (-0.94) & 62.42 (\textbf{0.55}) & 61.33 (-0.55) & 51.48 (-10.39) 
\\
$l_2, \epsilon=15$  & 47.73 & 46.41 (-1.33) & 48.20 (\textbf{0.47}) & 47.27 (-0.47) & 39.06 (-8.67)
\\
$l_2, \epsilon=20$  & 34.45 & 34.30 (-0.16) & 32.73 (-1.72) & 30.86 (-3.59) & 23.59 (-10.86) 
\\
$l_2, \epsilon=25$  & 21.64 & 24.30 (\textbf{2.66}) & 18.75 (-2.89) & 14.69 (-6.95) & 9.22 (-12.42) 
\\[.5ex]
$l_\infty, \epsilon=.125/255$\hphantom{m}  & 74.61 & 75.00 (\textbf{0.39}) & 75.23 (\textbf{0.62}) & 74.61 (0.00) & 72.19 (-2.42)
\\
$l_\infty, \epsilon=.25/255$  & 68.52 & 69.53 (\textbf{1.02}) & 69.53 (\textbf{1.02}) & 67.89 (-0.62) & 62.66 (-5.86) 
\\
$l_\infty, \epsilon=.5/255$  & 55.23  & 56.88 (\textbf{1.64}) & 56.25 (\textbf{1.02}) & 52.42 (-2.81) & 42.81 (-12.42) 
\\
$l_\infty, \epsilon=1/255$  & 34.77 & 37.73 (\textbf{2.50}) & 34.69 (-0.08) & 25.78 (-8.98) & 15.78 (-18.98) 
\\[0.5ex]

\end{tabularx}
\caption[Clean and robust accuracies for the ResNet18 and unstructured global magnitude pruning]{Clean and robust accuracies for all attack types and strengths for ResNet18 using unstructured global magnitude pruning with compression rates from 1 (no pruning) to 16. The relative margin (in \%) to the unpruned network is displayed in parenthesis. Bold numbers indicate increases in accuracy.}
\label{tab:clean_rob_accs}
\end{table*}

Note that, we treat the attack strength as discrete, which is only partially true as both, the B\&B $L_0$ and the C\&W $L_2$ attack are minimization attacks, and thus the attack strength can be viewed as being continuous. Minimization attacks return the minimal perturbation that leads to misclassification. 
Thus, we can retrospectively evaluate the robustness for any $\epsilon$. 
This is necessary to evaluate minimization attacks and fixed-epsilon attacks with the same metric. For each combination of model and attack, we choose a set of $\epsilon$-values such that the weakest attack decreases the accuracy only marginally
on the unpruned network and the strongest attack fools the network for more than 50\% of the images.

\subsection{Implementation}

We construct an evaluation pipeline in which
for each architecture and pruning method, a separate model with random weights is initialized. We add pruning masks to every layer of the model and set all mask-elements to one. The unpruned network is trained to convergence. Subsequently, we iterate over all compression rates and perform pruning, fine-tuning, and evaluation. 

We optimize the models with the ADAM algorithm \cite{kingma2014adam} over the categorical crossentropy loss with an initial learning rate of 0.001. We allow up to 150 epochs for training and implement early stopping observing the validation loss and patience of five epochs for CNN5 and 15 epochs for the other models. For the three bigger models, we apply dynamic learning rate scheduling, multiplying the learning rate by 0.3 after a patience period of twelve epochs. We run all experiments five times with different random seeds and report the average values. All experiments are implemented with Tensorflow 2.2.0 \cite{abadi2016tensorflow} and Foolbox 3.0.0~\cite{rauber2017foolbox}.

\section{Results} \label{sec:results}
In this section, we present the results of the experiments laid out above. For brevity, we only discuss the experiments we conducted on the ResNet18 architecture. However, we were able to identify the same properties we introduce in this section for the ResNet18 architecture for the other examined architectures. We provide the results for the CNN5, ResNet8, and the VGG11-like architectures in Appendix~\ref{appendix-res}.

\textbf{Clean Accuracies}
Table \ref{tab:clean_accuracies} shows the clean accuracies of the ResNet18 for all examined pruning methods and compression rates. Unsurprisingly, our results confirm the findings from previous work \cite{han2015learning} that NN pruning can enhance the clean accuracy for mild pruning and yields good results even for higher compression rates, specifically, the unstructured pruning approaches.

\textbf{Robustness Evaluation}
Figure \ref{fig:resnet18} shows the results of our experiments for the ResNet18. 
Interestingly the behavior is not consistent for all three attack types. When robustness is evaluated with the PGD attack (rightmost column), as done in most previous work, the expected behavior is observable: For moderate compression rates, the robustness remains stable with small in- and decreases depending on the pruning approach, while more extensive pruning does hurt the robustness considerably for all examined $\epsilon$-values. Evaluation against a B\&B $L_0$-adversary yields comparable results. In contrast, when evaluated with the C\&W $L_2$ attack (middle column), even for higher compression rates, a pruning approach exists that yields better robustness than the unpruned model. The filter-pruning methods (red lines) show consistently superior results where the magnitude-based approaches (dashed and dotted lines) appear to work better for moderate pruning, while pruning random filters (solid line) leads to better robustness for more extensive pruning.

Additionally, we can see that for every attack, there is a pruning approach that keeps the robust accuracy stable or increases it for moderate compression rates. So, if a defender has full information of the adversary with regards to the chosen attack type and strength, it is beneficial to optimize the model to that specific case. I.e., if it can be expected that the adversary will attack with an $L_2$ attack and a max $\epsilon$ of 15 (an assumption which might be derived from the nature of the problem) the model should be optimized for this specific case, and filter global magnitude pruning should be applied.
Such a level of knowledge about the adversary is unlikely, but motivates to look for sweet spots with favorable trade-offs for the defender regarding clean accuracy, robust accuracy, and the amount of the remaining weights.

\textbf{Sweet Spots}
Exemplary, we identify such sweet spots for the ResNet18 model when applying unstructured global magnitude pruning with a compression rate of two or four: Table \ref{tab:clean_rob_accs}
shows the absolute accuracies and margins to the unpruned model for all applied attacks and pruning ratios. For a compression rate of two we see a minimal decrease of up to 1.33\% in robust accuracy compared to the unpruned model, when attacked with the $L_2$ attack, while for all other combinations of attack type and strength, the robustness is increased by a margin of 3.36\%. Even a compression rate of four does not reduce the robustness for any attack examined by more than 2.89\%, while for some scenarios, we see an increase of the respective robustness by up to 1.25\%. We can find these sweet spots for the remaining architectures.

\section{Conclusion}
\label{sec:conclusion}
In this work, we shed some light on the impact of NN pruning in the face of adversaries. We conducted an extensive series of experiments with an ensemble of pruning approaches and attack methods that were carefully selected to provide a broad view.

Small increases in robustness for mild pruning were already noticed in small-scaled experiments in prior work and we confirm this for a wide variety of attack-types, attack strengths, pruning approaches, and compression rates. The stronger increase in robustness against an $L_2$ adversary, observable in the middle column of Figure \ref{fig:resnet18}, is intriguing and calls for further research. 

An intuition why robustness might increase with pruning and thus, contradicting the general assumption that capacity helps \cite{madry2018towards}, could be the following: \citet{szegedy2013intriguing} and \citet{goodfellow2014explaining} argue that adversarial examples leverage so-called \textit{blind spots}, which are low density regions of the training data distribution. Pruning aims to eliminate the least important parts of a DNN, and for smaller compression rates, the parts removed contain proportionally more of these so called \textit{blind spots}.

Our results validate that the method and extent of NN pruning open up additional possible strategies for adversary-aware deep learning practitioners. Furthermore, we show that by making additional assumptions about potential adversaries, we can identify optimal pruning strategies. Factors to consider are possible attack types and strengths. 

Practitioners should not only think about NN pruning for applications operating under computational and memory constraints, but our findings suggest that security-sensitive use cases might benefit from a carefully selected pruning strategy. NN pruning can simultaneously increase a model's clean accuracy and its robustness against a wide variety of adversarial attack methods and strengths. This is valid for both cases: When resources are not a limiting factor and under resource constraints.

Our choice of perturbation budgets follows a simple heuristic. While we can assert that an $L_0$ and an $L_\infty$ attack are similarly successful in fooling a model, we cannot make any statements about the real strength of an attack. However, the limitations of the $L_p$ norms are well known and discussed in the adversarial machine learning community \cite{carlini2019evaluating}.

While we deliberately refrained from incorporating adversarial training methods due to its negative impact on the clean accuracy, future work should examine if a combination of mild pruning, moderate adversarial training, and fine-tuning leads to a significant rise in adversarial robustness while not hurting the clean accuracy compared with a naturally trained, unpruned network. A combination achieving this, is considered to be the, so far unreached, ``holy grail'' of adversarial machine learning research. 

\printbibliography

\appendix

\section{Attack Implementation Details}
\label{appendix-attacks}
For the CNN5, ResNet8 and VGG11-like architectures we evaluate the robustness on 1000 images from the test set, for the ResNet18 we use a set of 256 images. For the PGD attack we set the relative step size to .1/3, the value of $\epsilon$, and allowed 40 steps with random start. 
For the C\&W $L_2$ attack, we perform 9 binary search steps with 5000 steps each, we found a step size of 1 and an initial cost of 100 deliver the best results. The confidence parameter is 0.
For the B\&B $L_0$ attack, we sample the starting points from our test set. We allow up to 30 binary search steps, with 500 steps each. We choose a initial learning rate of 1e7 and decay the learning rate 30 times by .5.

\section{Results for CNN5, ResNet8 and VGG11-like}
\label{appendix-res}
Here (on the next page) we report the cumulated results of our experiments for the architectures CNN5 in Subfigure~\ref{fig:cnn5}, ResNet8 in Subfigure~\ref{fig:resnet8}, and VGG11-like in Subfigure~\ref{fig:vgg}. They confirm the results reported above for ResNet18.

\begin{figure*}
    \begin{subfigure}{\textwidth}
        \centering
        \resizebox{.9\textwidth}{!}{\input{tikz-figs/cnn.tex}}
        \subcaption{CNN5 architecture}
        \label{fig:cnn5}
    \end{subfigure}
    \begin{subfigure}{\textwidth}
        \centering
        \resizebox{.9\textwidth}{!}{\input{tikz-figs/resnet8.tex}}
        \subcaption{ResNet8 architecture}
        \label{fig:resnet8}
    \end{subfigure}
\end{figure*}
\begin{figure*}[t]

        \centering
        \resizebox{.9\textwidth}{!}{\input{tikz-figs/vgg.tex}}
        \subcaption{VGG11-like architecture}
        \label{fig:vgg}
    
        \caption[]{Adversarial robustness against the B\&B $L_0$ (\textbf{left} column), C\&W $L_2$ (\textbf{middle}) and PGD $L_\infty$ attacks (\textbf{right}). Each line in each plot depicts a pruning method over the compression rates from one to 64. (Best viewed in color.)}
    \label{fig:all-res}
\end{figure*}%

\vfill

\end{document}